\documentclass[letterpaper, 10 pt, conference]{ieeeconf}  
\IEEEoverridecommandlockouts 
\makeatletter
\let\NAT@parse\undefined
\makeatother

\usepackage[pdftex]{graphicx}
\usepackage{amsmath} 
\usepackage{amssymb}
\usepackage{subfigure}
\usepackage{multirow}
\usepackage{array,booktabs}
\usepackage{diagbox}
\usepackage{balance}
\usepackage{mathtools}
\usepackage{subfigure}
\usepackage{verbatim}
\usepackage{adjustbox}
\usepackage{algorithm}
\usepackage{algpseudocode}
\usepackage{subcaption}
\usepackage{romannum}
\usepackage{hyperref}
\usepackage{xurl}
\usepackage[labelsep=period]{caption}
\usepackage[flushleft]{threeparttable}
\usepackage[table,xcdraw,dvipsnames]{xcolor}
\usepackage{tablefootnote}
\usepackage{multirow}   % for multirow cells
\usepackage{multicol}   % for multicols (rarely used in tables but good to have)
\usepackage{array}      % for better column formatting
\usepackage{booktabs}   % for cleaner horizontal rules (\toprule, \midrule, \bottomrule)
\usepackage{tabularx}   % in preamble
\hypersetup{
    colorlinks=true,
    linkcolor=black,
    citecolor=black,
    filecolor=black,
    urlcolor=blue,
}
\usepackage[table]{xcolor}
\usepackage[numbers]{natbib}
\usepackage{soul,color}
\usepackage{multirow}
\usepackage{lipsum}
\usepackage{makecell}

\usepackage{tikz}

\usepackage{user_math}
\usepackage{user_misc}
\usepackage{user_acronyms}
\usepackage{user_SIunits}

\begin{document}

\title{\LARGE \bf Kitchen Robotic Manipulation utilizing Foundation Models}

\author{Myung-Hwan Jeon${}^{1*}$, Sankalp Yamsani${}^{2}$, and Joohyung Kim${}^{2}$
\thanks{$^\dagger$This work was supported by the National Research Foundation of Korea(NRF) grant funded by the Korea government(MSIT) (RS-2024-00337118) and the regional innovation system \& education (RISE)-(Regional Growth Innovation LAB) program through the Gyeongbuk RISE Center, funded by the Ministry of Education (MOE) and the Gyeongsangbuk-do, Republic of Korea(2025-rise-15-105).}
\thanks{$^{1}$ M. Jeon is with the School of Electronic Engineering, Kumoh National Institute of Technology, Gyeongbuk, S. Korea. {\tt\small mhjeon@kumoh.ac.kr}}
\thanks{$^{2}$S. Yamsani and J. Kim are with the Kinetic Intelligent Machine Lab (KIMLAB), University of Illinois Urbana-Champaign, Champaign, IL 61801 USA. {\tt\small [yamsani2, joohyung]@illinois.edu}}%
}

\maketitle
\thispagestyle{empty}
\pagestyle{empty}

\begin{abstract}

Deploying robots in everyday human environments requires perception systems that are both robust and adaptable to diverse, dynamic conditions. In this work, we present a modular perception pipeline for household manipulation tasks, with a focus on dishware handling in kitchen environments. The pipeline integrates open-vocabulary object detection, multi-view segmentation, instance-aware 3D reconstruction, and a 2D-3D feature fusion strategy for 6D pose estimation and grasp planning. Its modular design enables systematic substitution of multiple visual and geometric foundation models, allowing us to identify the best-performing configuration through extensive evaluation on a custom kitchen dataset. The best-performing configuration (LLMDet + SAMv2 + DINOv2 + GeoTransformer) achieves an ADI of 89.12\% on the 20-scene kitchen benchmark with cluttered and occluded conditions. Furthermore, real-world demonstrations confirm that the best configuration can be deployed on physical robots without environment-specific retraining, successfully executing tasks such as sink-to-dishwasher transfer and cup stacking. It validates the adaptability and scalability of the pipeline and highlights its potential as a practical framework for household robotic systems. Our code and supplementary materials are available at \href{https://raivlab.github.io/FM_kitchen/}{https://raivlab.github.io/FM\_kitchen/}.

\end{abstract}

\section{Introduction}\label{sec1}
With the rapid advancement of robotic technologies, the integration of robots into everyday human environments has garnered increasing attention. From assistive robotic systems in kitchens \cite{tri_kitchen, fukuzawa2021clean} to autonomous cleaning \cite{mega2025clean} and general-purpose household manipulation \cite{gu2025humanoid}, researchers have been devoted to developing robots capable of operating effectively in real-world settings. However, deploying such systems in human environments remains a significant challenge, particularly due to the need for adaptability and robust perception in unstructured and dynamic contexts. 

%FIGURE
\begin{figure}[!t]
	\centering
	\includegraphics[width=0.99\columnwidth]{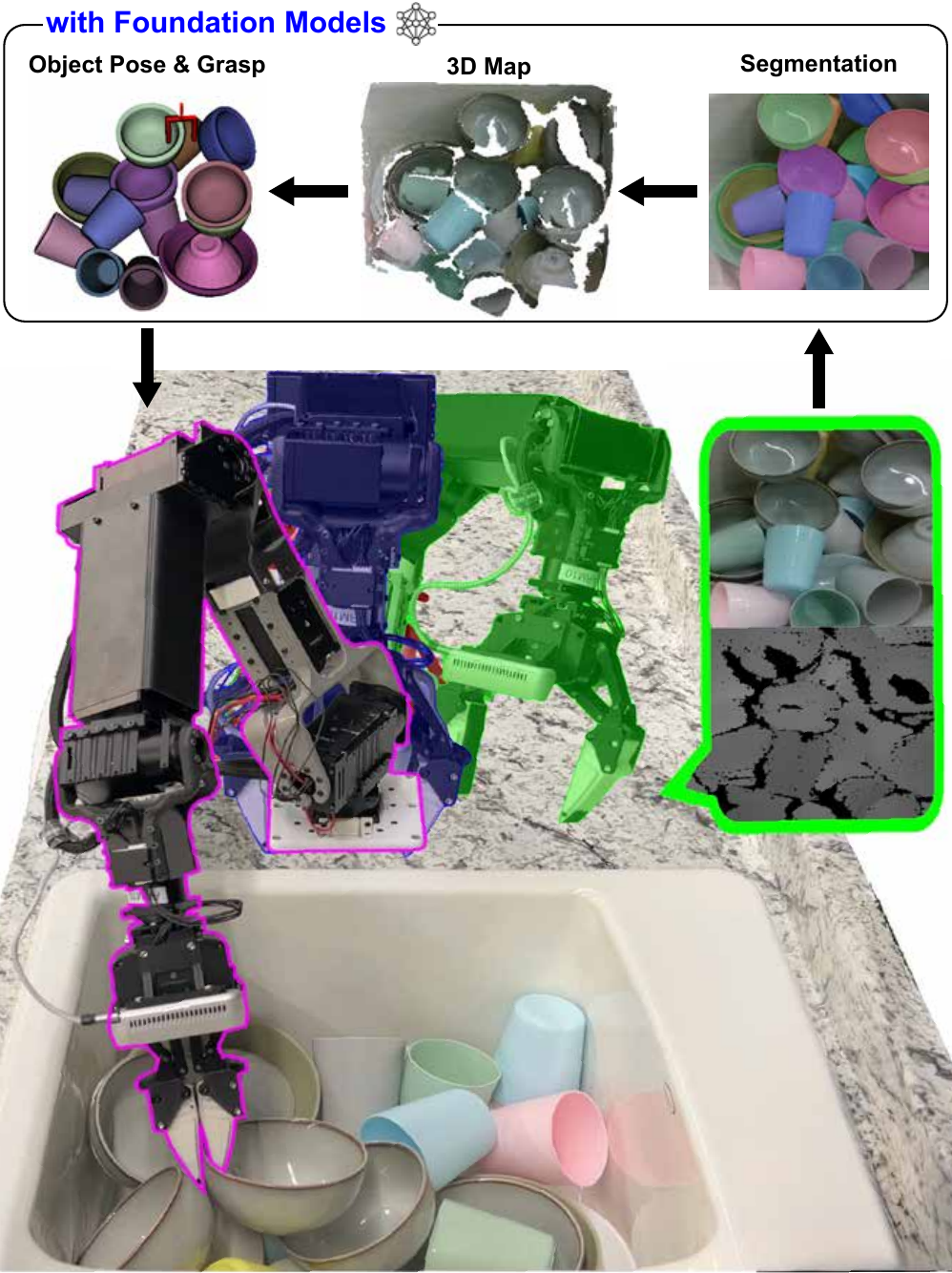}	
  	\caption{The robot manipulator scans the workspace and processes the collected data using a proposed perception pipeline built on multiple foundation models. Based on these results, it estimates the pose and grasp configurations of the dishware, enabling the system to manipulate the items and demonstrate generalizable perception in kitchen environments.
	}
	\label{fig:overview}	
\end{figure}

Recent developments in robotic systems such as plug-and-play robotic systems \cite{kim2023papras} and the emergence of affordable robots \cite{unitree} highlight ongoing efforts to lower deployment barriers by reducing hardware costs, installation overhead, and integration complexity. Nevertheless, achieving true scalability in real-world environments critically depends on perception pipelines that can adapt to diverse and dynamic settings without the need for extensive customization or environment-specific training.

Human environments, in particular, demand robots that can operate under constantly changing conditions and handle unfamiliar surroundings with minimal manual intervention. Traditional vision-based robotic systems depend heavily on task-specific datasets and supervised learning, which often fail to generalize beyond predefined environments. This limitation restricts their usability in real-world scenarios. In contrast, recent advancements in deep learning have led to the emergence of foundation models \cite{oquab2023dinov2, ravi2024sam, naver2024dust3r, liu2023grounding}, which demonstrate remarkable generalization capabilities across different domains. Unlike conventional models, foundation models leverage large-scale pretraining and self-supervised learning, enabling robots to recognize and interact with their surroundings without requiring extensive retraining. These properties make foundation models highly suitable for robotic systems, supporting robust perception and adaptability across diverse and dynamic environments.

These advances underscore the growing potential of foundation models for robotics. A critical question still remains: how can such models be systematically integrated into perception pipelines that must operate reliably in everyday human environments? Although foundation models demonstrate strong generalization, their practical deployment in robotics requires bridging high-level perception with the concrete demands of manipulation. Meeting this challenge requires frameworks that can flexibly combine diverse models and adapt across tasks and environments, thereby enabling robust and scalable performance in real-world environments.

In our work, we push the potential of robotic systems by constituting a unified perception pipeline to address a key household task: recognizing and manipulating dishware in the sink. This pipeline actively leverages multiple foundation models to operate without environment-specific finetuning and follows a modular architecture in which each component is designed to be flexibly swappable. This design allows systematic exploration of diverse foundation model combinations to identify optimal configurations. It yields a robust and competitive system for kitchen object recognition without retraining, while ensuring precise manipulation in dynamic kitchen environments. To demonstrate scalability, we validate the pipeline across multiple robotic configurations and kitchen environments. Through extensive evaluation, we exemplify that the constituted pipeline successfully adapts to various robot systems and new environments, highlighting its potential for scalable and practical household automation.

Differing from previous methods, our method presents the following contributions:

\begin{itemize}

    \item We present a system-level integration of multiple foundation models into a unified 6D-pose perception pipeline that operates across diverse kitchen environments without environment-specific training.
    
    \item The integrated system is composed of modular components, each of which can be flexibly swapped with alternative models. To validate this design, we systematically evaluate 24 combinations of visual and geometric foundation models on a real-world kitchen dataset.
    
    \item We demonstrate the effectiveness of the integrated system through experiments on several robotic platforms performing kitchen manipulation tasks. The integrated system adapts to varied human environments without further tuning, highlighting the scalability and practical benefits of our solution for household automation tasks.

\end{itemize}

\section{Related Works}
\label{sec:related}

\subsection{Foundation Models}
\label{sec:related_1}

Foundation models trained on large-scale datasets have demonstrated remarkable generalization across diverse modalities. DINOv2 \cite{oquab2023dinov2} and BEiT \cite{beit} learn rich image representations without task-specific supervision, providing transferable features that are robust across downstream tasks. Extending beyond unimodal vision, CLIP \cite{radford2021clip} aligns visual and textual embeddings through joint training on image–text pairs, enabling open-set recognition and language-guided reasoning. These advances have further motivated open-world object detection \cite{liu2023grounding, fu2025llmdet}, where vision–language grounding is combined with large-scale language modeling to support open-vocabulary detection of novel objects.

For 3D data processing, specialized architectures have been introduced to integrate geometric priors. GeDi \cite{poiesi2022gedi} extracts geometry-aware descriptors that capture local 3D features for robust registration and matching, while BufferX \cite{seo2025buffer} enables efficient large-scale point cloud handling for practical deployment. GeoTransformer \cite{qin2023geotransformer} further incorporates geometric structures into transformer attention mechanisms, improving global context modeling and enhancing point cloud registration performance.

%FIGURE
\begin{figure*}[!t]
	\centering
	\includegraphics[width=0.99\textwidth]{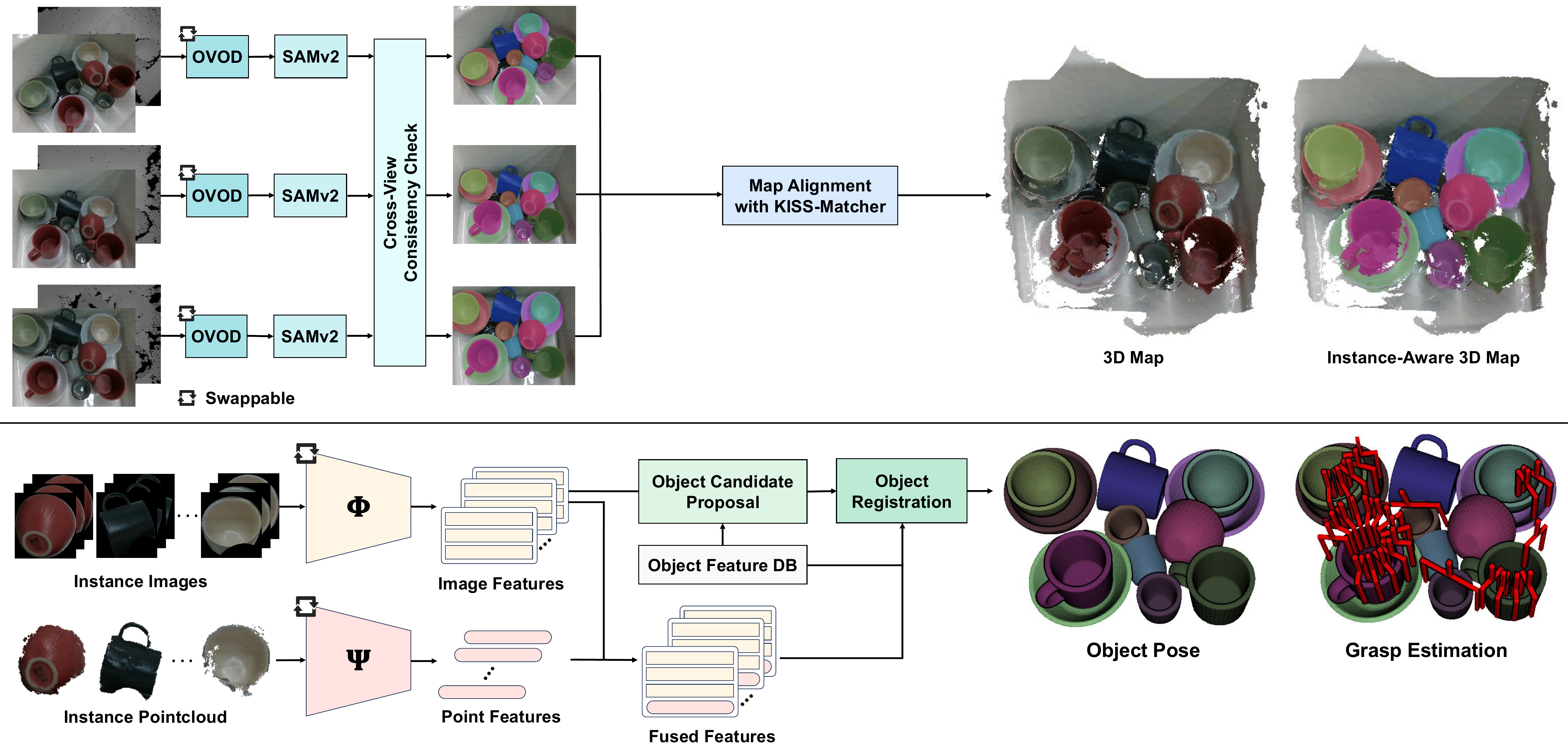}	
  	\caption{Overview of the perception pipeline. The system integrates \ac{OVOD} methods with SAMv2 for multi-view consistent segmentation, followed by instance-aware 3D map reconstruction. A 2D–3D feature fusion strategy combines image features and point features to enable object candidate proposal and registration. The resulting object poses are then used for grasp estimation, with all modules designed to be swappable to validate diverse combinations of visual foundation models $\Phi$ and geometric foundation models $\Psi$.
	}
	\label{fig:pipeline}	
\end{figure*}
%FIGURE

\subsection{Foundation Models for Robotic Perception}
\label{sec:related_2}

Recent developments have extended the foundation model paradigm to a wide range of robotic perception tasks. For object recognition, \ac{SAM} \cite{ravi2024sam} demonstrates accurate and efficient image segmentation across diverse scenes, while approaches such as SAM-6D \cite{lin2024sam}, CNOS \cite{nguyen2023cnos}, and FoundationPose \cite{wen2024foundationpose} introduce scalable pipelines for object recognition and 6D pose estimation. In addition, FoundationGrasp \cite{Tang2025FoundationGrasp} provides generalizable grasp detection, a capability essential for robotic manipulation. While direct grasp synthesis methods such as GraspAnything \cite{vuong2024grasp} can handle simple grasping tasks without explicit pose estimation, the manipulation scenarios, such as placing dishes into dishwasher racks or stacking cups, require accurate 6D pose knowledge to enable subsequent manipulation steps with precise spatial constraints.

For 3D scene understanding, foundation models have also shown strong potential. Depth Anything \cite{yang2024depth} achieves robust monocular depth estimation across varied domains, while VFMM3D \cite{ding2024vfmm3d} adapts the foundation model paradigm to 3D object detection, enabling spatial reasoning in cluttered environments. Furthermore, models such as DUSt3R \cite{naver2024dust3r} and MASt3R-SLAM \cite{murai2025mast3r} enable dense geometric reconstruction from multi-view images or sequential observations, thereby supporting consistent large-scale scene understanding.

Several recent works integrate foundation models into mobile robotic systems for perception. In \cite{bd_spot}, visual foundation models serve as a scene-understanding front-end for a robot system, providing semantic categories that drive obstacle and hazard reasoning during navigation in dynamic environments. In \cite{zhang2024inno}, the \ac{SAM} \cite{ravi2024sam} is paired with a robotic arm to produce object masks that are converted into real-time grasping commands via depth-based geometry. Both works demonstrate that pretrained foundation models can replace task-specific perception modules in robotic systems. Our work extends this line of integration to inventory-based kitchen manipulation. We combine multiple visual and geometric foundation models to perceive the environment and target objects, ensuring generalizable performance for robot manipulation without the need for environment-specific training.

A growing body of work composes foundation models for robotic manipulation. For household pick-and-place, OK-Robot \cite{liu2024ok} combines an open-vocabulary detector with pretrained AnyGrasp \cite{fang2023anygrasp} grasping over a pre-recorded scan of each home. TidyBot \cite{wu2023tidybot} uses an \ac{LLM} to infer personalized object-placement rules. Both rely on a per-environment scan or per-user rules. VoxPoser \cite{huang2023voxposer} takes a different route. A LLM writes code that queries vision-language models, composing 3D value maps for a model-based motion planner. A separate line of work trains end-to-end vision-language-action policies that map images and language directly to actions. RT-2 \cite{Brohan2023rt2} co-fine-tunes a vision-language model on robot action together with web-scale VQA data, producing robot actions as language tokens. OpenVLA \cite{kim2024openvla} trains a 7B-parameter VLA on the Open X-Embodiment dataset \cite{embodimentcollaboration2025openxembodimentroboticlearning}. Octo \cite{team2024octo} trains a transformer-based diffusion policy on the same dataset. These end-to-end policies achieve impressive generalization across tasks and robot embodiments. However, they fold perception and control into a single learned model. Such a model cannot be reused with a different motion planner or sensor configuration without retraining. Our work targets a different design point. We use a kitchen-scoped pipeline with a shared CAD inventory. This pipeline produces explicit 6D poses for any classical motion planner, performs no task-specific finetuning, and keeps perception and control as separate modules.

\section{Method}
\label{sec:method}

\subsection{Notation}
\label{sec:method_notation}

We use the following notation to describe the proposed method. The
transformation from coordinate system $A$ to coordinate system $B$ is denoted
by ${}^{A}_{B}T$. It consists of a rotation $R_{AB} \in SO(3)$ and a
translation $t_{AB} \in \mathbb{R}^{3}$. The symbol $W$ denotes the world
coordinate system. The subscript $(\cdot)_i$ identifies the index of the
corresponding variable. The subscript $(\cdot)_{1:n}$ denotes the group of
values from the initial value through the $n^{\mathrm{th}}$ value. The
operator $[\cdot \mathbin\Vert \cdot]$ denotes concatenation, and
$\lVert\cdot\rVert_2$ denotes the L2 norm.

\subsection{Overview}
\label{sec:method_Overview}

We focus on an inventory-based kitchen manipulation task, with particular emphasis on recognizing dishware. Items such as plates, cups, and bowls fall into a limited number of categories and are often part of standardized sets, making kitchen environments generally predictable. As a result, these objects are rarely replaced or supplemented with new ones, leading to a highly stable inventory. Given these characteristics of kitchen environments, we assume that 3D CAD models of the target objects are available and that the object inventory is known, allowing us to constrain the recognition space. The CAD models allow us to use predefined stable grasp configurations that have been prevalidated on each object type. These configurations ensure higher manipulation success rates in precision tasks such as dishwasher placement and cup stacking.

Building on this setup, we describe the perception-and-manipulation pipeline as follows. As shown in \figref{fig:overview}, given a manipulator equipped with an RGB-D camera that follows precomputed end-effector waypoints, the robotic system sequentially captures RGB-D images along with the corresponding proprioceptive state of the manipulator at the waypoint. Taking this data as input, we aim to recognize the target objects, estimate 6D poses, and generate grasp candidates for robot manipulation. Based on these estimates, the robot system executes the pick and place of the dishware from the sink.

\subsection{Perception System}
\label{sec:method_perception}

As shown in \figref{fig:pipeline}, the perception pipeline contains multiple
steps that run concurrently with workspace scanning. This scanning produces
$n$ RGB-D images $I^{\mathrm{rgb}}_{1:n}$ and
$I^{\mathrm{depth}}_{1:n}$, together with the corresponding camera poses
$C_{1:n}$.

\begin{align}
I^{\mathrm{rgb}}_{1:n}
&\in \mathbb{R}^{H \times W \times 3},\qquad
I^{\mathrm{depth}}_{1:n} \in \mathbb{R}^{H \times W},
\label{eq:rgbd-inputs}\\
C_{1:n}
&\triangleq \left\{{}^{B}_{C_i}T \in SE(3)\right\}.
\label{eq:camera-poses}
\end{align}

The camera is attached to the end-effector of the robot manipulator. Using
these inputs, the pipeline estimates the object's 6D pose ${}^{B}_{O}T$ and
$k$ grasp candidates $G_{1:k}$.

\begin{equation}
G_{1:k}
\triangleq \left\{{}^{B}_{G_i}T \in SE(3)\right\}.
\label{eq:grasp-candidates}
\end{equation}

These estimates are based on the relative transformation between the robot
system's base coordinate frame $B$ and the object coordinate frame $O$.

\subsubsection{Instance Segmentation across multi-view images}
\label{sec:method_segmentation}

Effective object recognition requires consistent segmentation across multi-view images. To obtain it, we combine \acf{OVOD} with SAM. In each image, \ac{OVOD} generates bounding-box prompts for \ac{SAM}. Cross-view verification then produces consistent object masks.

Given the image set $I^{\mathrm{rgb}}_{1:n}$, we first detect the target objects using the simple text prompt (\texttt{"dishware"}). The detected bounding boxes are then used as segmentation prompts. This produces the instance masks $M_{1:n,1:m}$, where $m$ denotes the number of instances per image. Because the masks are generated independently for each image, they do not consistently correspond to the same physical object across different views. To resolve this inconsistency, we establish cross-view correspondences by projecting masks from one view onto the others using the associated depth images and camera poses. We then compute the \ac{IoU} between the projected and detected masks to measure their overlap. If the \ac{IoU} exceeds a predefined threshold, the corresponding masks are assigned the same object ID. We perform this verification through a bidirectional cross-check to ensure robust and consistent instance associations across all views.

\subsubsection{Instance-Aware 3D Map Reconstruction}
\label{sec:method_map}

To enable accurate object recognition, we first construct an instance-aware 3D
semantic map. Its inputs are the RGB-D image set $\{I^{\mathrm{rgb}}_{1:n}, I^{\mathrm{depth}}_{1:n}\}$, the instance masks $M_{1:n,1:m}$, and the camera poses $C_{1:n}$. When the camera poses are sufficiently accurate, the RGB-D frames can be reliably fused into a point cloud map using \ac{ICP} refinement. In practice, however, the quality of the robot manipulator and its calibration
can introduce uncertainty. The resulting pose errors may misalign the map. To
address this issue, we first apply coarse 3D point cloud registration based on
feature matching \cite{lim2025icra-KISSMatcher}. We then use ICP refinement for precise alignment.

After establishing geometric alignment, we augment the 3D map with semantic information by assigning instance IDs from the masks $M_{1:n,1:m}$. The resulting 3D semantic map ${}^{B}\mathrm{Map}$ is represented as a set of points. Each point is defined by the four-dimensional tuple $(x,y,z,\mathrm{id})$.

\begin{align}
\label{eq:aligned_point_cloud}
^{B}\mathrm{Map} = \{ (x, y, z, \text{id}) \mid (x, y, z) \in \mathbb{R}^3,\, \text{id} \in \mathbb{N} \}
\end{align}

\subsubsection{Object Classification and 6D Pose Estimation}
\label{sec:method_pose}

Our objective is to determine the class and 6D pose of each segmented instance in the 3D map. We first introduce a feature fusion strategy that integrates image and point features, inspired by \cite{caraffa2024freeze}. Building on this fused representation, estimation proceeds through three sequential stages: (1) database preparation, (2) object candidate proposal, and (3) object registration. This design allows us to leverage rich multimodal features while maintaining a clear and modular pipeline for classification and pose estimation.

\begin{figure}[t]
    \begin{algorithm}[H]

        \algrenewcommand\alglinenumber[1]{#1:}
        \caption{2D-3D Feature Fusion for a Single Instance}
        \label{alg:fusion_single}
        
        \hspace*{\algorithmicindent} \textbf{Input:} \\
        \hspace*{\algorithmicindent} RGB--D frames $\{I^{rgb}_{1:n}, I^{depth}_{1:n}\}$ \\
        \hspace*{\algorithmicindent} instance masks $M_{1:n,k}$ \\
        \hspace*{\algorithmicindent} camera poses $\mathcal{C}_{1:n}$ \\
        \hspace*{\algorithmicindent} intrinsics $K$ \\
        \hspace*{\algorithmicindent} voxel size $v$ \\
        \hspace*{\algorithmicindent} \textbf{Output:} fused feature representation $\mathcal{F}$
        
        \begin{algorithmic}[1]
            \State $\mathcal{F}^I \gets \emptyset$, $P \gets \emptyset$
            \For{each frame $i=1..n$}        
                \State $I^{rgb'}_{i}, I^{depth'}_{i}, K' = \mathrm{PreProcess}(I^{rgb}_i, I^{depth}_i, M_{i,k}, K)$
                \Statex \Comment{background removal + crop + resize, update $K$}
                \State $f^I_{i,\mathrm{cls}}, f^I_{i,\mathrm{patch}} = \Phi(I^{rgb'}_i)$         
                \State Resize $f^I_{i,\mathrm{patch}}$ to $224\times224$
                \State $\tilde f^I_i \gets \Call{PCA}{f^I_{i,\mathrm{patch}}}$
                \State $P_i = \mathrm{ToPointCloud}(I^{depth'}_i, K', \mathcal{C}_i)$
                \State $\mathcal{F}^I = \mathcal{F}^I \cup \tilde f^I_i$, \quad $P \gets P \cup P_i$
            \EndFor
            \State $(\hat{\mathcal{F}}^I, P_v) = \mathrm{VoxelPool}(\mathcal{F}^I, P, v)$
            \Statex \Comment{aggregate per voxel}
            \State $\mathcal{F}^P \gets \Psi(P_v)$
            \State $\mathcal{F} = \Big[\, \tfrac{\hat{\mathcal{F}}^I}{\|\hat{\mathcal{F}}^I\|_2} \;\Big\Vert\; \tfrac{\mathcal{F}^P}{\|\mathcal{F}^P\|_2} \,\Big]$
            \Statex \Comment{Normalize and fuse}
            \State \Return $\mathcal{F}$
        \end{algorithmic}
    \end{algorithm}
\end{figure}

\paragraph{2D and 3D Feature Fusion}

For each segmented object instance, we construct a fused representation in
four steps. (1) Given the RGB-D frames $\{I^{\mathrm{rgb}}_{1:n}, I^{\mathrm{depth}}_{1:n}\}$, camera poses $C_{1:n}$, and instance masks $M_{1:n,1:m}$, a visual foundation model $\Phi$ extracts class and patch tokens from the cropped object regions. (2) We project the patch tokens into 3D space using the corresponding depth frame and camera pose. The spatial resolution of the patch tokens differs from that of the underlying point cloud. We therefore aggregate the back-projected features through voxel pooling to obtain per-point image features. (3) In parallel, a geometric foundation model $\Psi$ computes point-level features on the object point cloud. (4) We L2-normalize and concatenate the two feature streams, as shown in \equref{eq:fused_feat}. This operation forms the unified per-point feature representation used by the downstream classification and pose-estimation stages.

\begin{align}
\label{eq:fused_feat}
\mathcal{F} = \Big[\, \tfrac{\hat{\mathcal{F}}^I}{\|\hat{\mathcal{F}}^I\|_2} \;\Big\Vert\; \tfrac{\mathcal{F}^P}{\|\mathcal{F}^P\|_2} \,\Big],
\end{align}

Here, $\widehat{\mathcal{F}}^{I}$ and $\mathcal{F}^{P}$ denote the projected
image features and point-level features, respectively. We use simple
concatenation after L2 normalization for feature fusion. This approach
preserves information from both modalities without requiring additional
training. More sophisticated fusion strategies, such as attention-based mechanisms or learned projections, could potentially improve performance. However, they would require additional training and model fine-tuning, which would reduce practical applicability. The normalization step ensures that both feature modalities contribute equally to the fused representation. Our experimental results demonstrate that this simple approach effectively combines complementary information from visual and geometric features. The fusion strategy captures both geometric structure and semantic cues. It therefore provides a robust representation for downstream object classification and pose estimation. Further details are given in \algoref{alg:fusion_single}.

\paragraph{Database Preparation}

We prepare a feature database by rendering multiple synthetic views of each
CAD model. We then extract the corresponding image and point features. For
every model, we store (i) class-level tokens, (ii) patch-level
tokens, and (iii) fused features $(P_{\mathrm{ref}},\mathcal{F}_{\mathrm{ref}})$.
Here, $P_{\mathrm{ref}}$ denotes the model points, and $\mathcal{F}_{\mathrm{ref}}$ denotes their associated features. This database
enables fast object candidate proposal and registration against query
instances observed in real scenes.

\paragraph{Object Candidate proposal}

For a query instance, we form the image feature representation by concatenating the class and patch tokens, $[f^{I}_{\mathrm{cls}} \mathbin\Vert f^{I}_{\mathrm{patch}}]$. We compute cosine similarities between this representation and the tokens of each CAD model stored in the database. For each model, we aggregate these similarities to produce a candidate score. We then select the top-$K$ models whose scores exceed a predefined threshold as potential matches. This step effectively prunes the search space, leaving only plausible object candidates for subsequent geometric alignment.

\paragraph{Object registration}

Object registration proceeds in four steps.
(1) For the plausible object candidates, we perform coarse alignment between the query point cloud $P_{\mathrm{query}}$ and the model points $P_{\mathrm{ref}}$. The alignment uses correspondence-based registration in the fused feature space $(\mathcal{F}_{\mathrm{query}},\mathcal{F}_{\mathrm{ref}})$.
(2) We evaluate each alignment using the bidirectional overlap ratios defined by the equations below.
(3) We discard candidates with insufficient correspondences or low overlap. We assign a preliminary score as the average of the two overlap ratios.
(4) For the best candidate for each instance, we further refine the
pose by running ICP between the transformed model point cloud and the
reconstructed 3D map. This refinement reduces the remaining misalignment after
coarse registration.

\begin{align}
\label{eq:or_ref2query}
\mathrm{OR}_{\text{ref}\to \text{query}} &= \mathcal{N}_\epsilon(P \mid Q_m), \\
\label{eq:or_query2ref}
\mathrm{OR}_{\text{query}\to \text{ref}} &= \mathcal{N}_\epsilon(Q_m \mid P), \\
\label{eq:N_def}
\mathcal{N}_\epsilon(X \mid Y) &= 
\frac{1}{|X|} \sum_{x \in X} \mathbf{1}\!\left[\, \min_{y \in Y} \|x-y\|_2 < \epsilon \,\right],
\end{align}

, where $\mathcal{N}_\epsilon(X \mid Y)$ measures the fraction of points in $X$ within a distance threshold $\epsilon$ of any point in $Y$. Here, $\mathbf{1}[\cdot]$ denotes the indicator function, and $\epsilon$ is set to the voxel size used in point cloud processing.

\subsubsection{Grasp Estimation for Robot Manipulation}
\label{sec:method_grasp}

We estimate grasps for manipulating the target objects. Because 3D CAD models
of all target objects are available, we predefine grasp candidates for each
object. These candidates can be adapted to the gripper type of the robot
manipulator. We then refine them through a series of feasibility checks so that
only physically accessible grasps remain.

First, we discard objects that are severely occluded by other objects. For each
remaining accessible object, we generate grasp candidates from its estimated
6D pose. We evaluate each candidate for reachability using inverse kinematics
and reject candidates without valid solutions. We also remove candidates that
could result in grasping multiple objects simultaneously. Finally, motion
planning eliminates candidates that would cause collisions with the
environment or require infeasible manipulator motions. This multi-stage
filtering process yields feasible grasp configurations that are both
kinematically valid and physically executable.

\begin{figure}[!t]

    \centering
    \begin{minipage}{0.99\columnwidth} 
    \centering
    \subfigure[First Kitchen]
    {%
    \includegraphics[width=0.44\columnwidth]{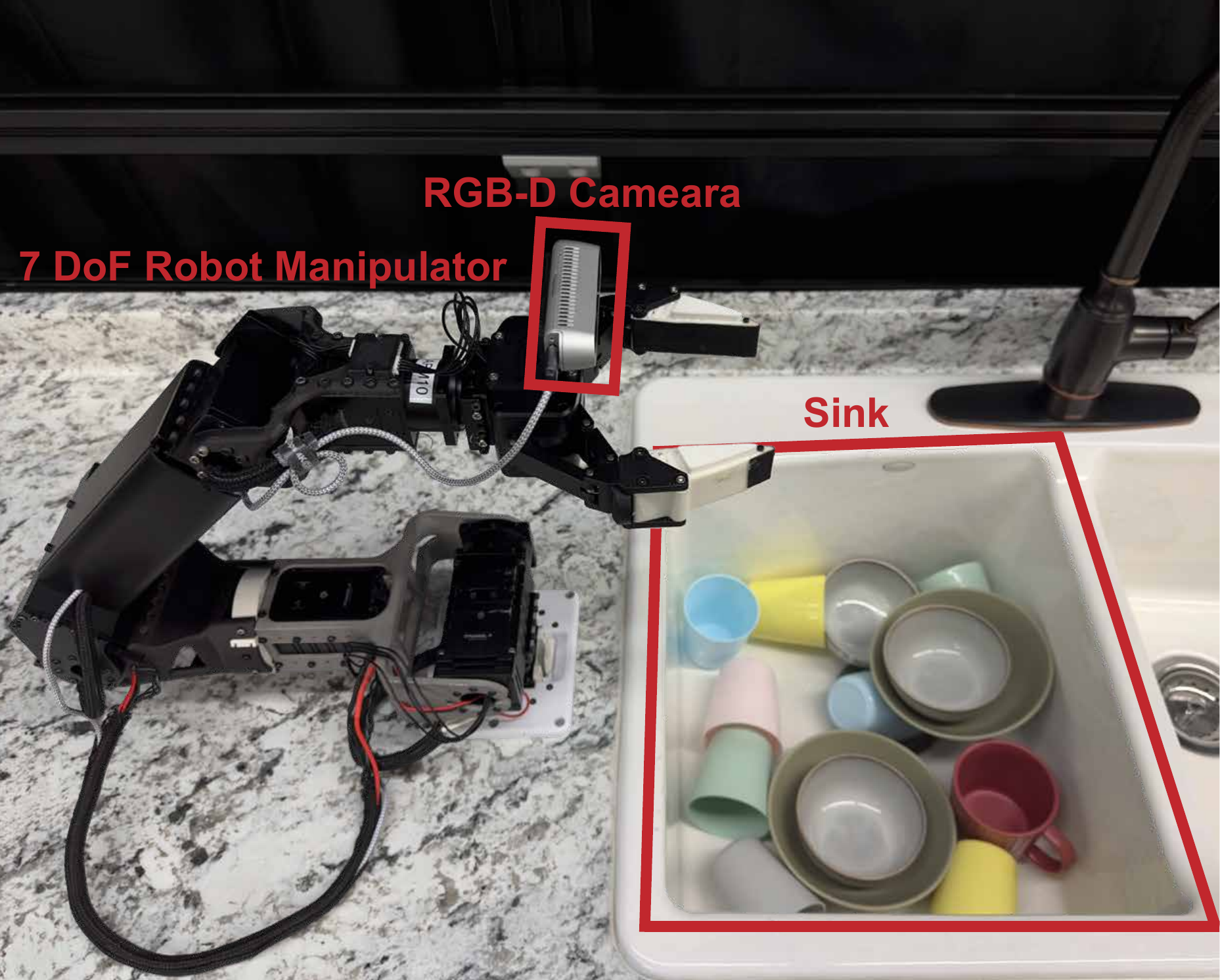}
    \label{fig:system1}
    }
    \subfigure[Second Kitchen]
    {%
    \includegraphics[width=0.44\columnwidth]{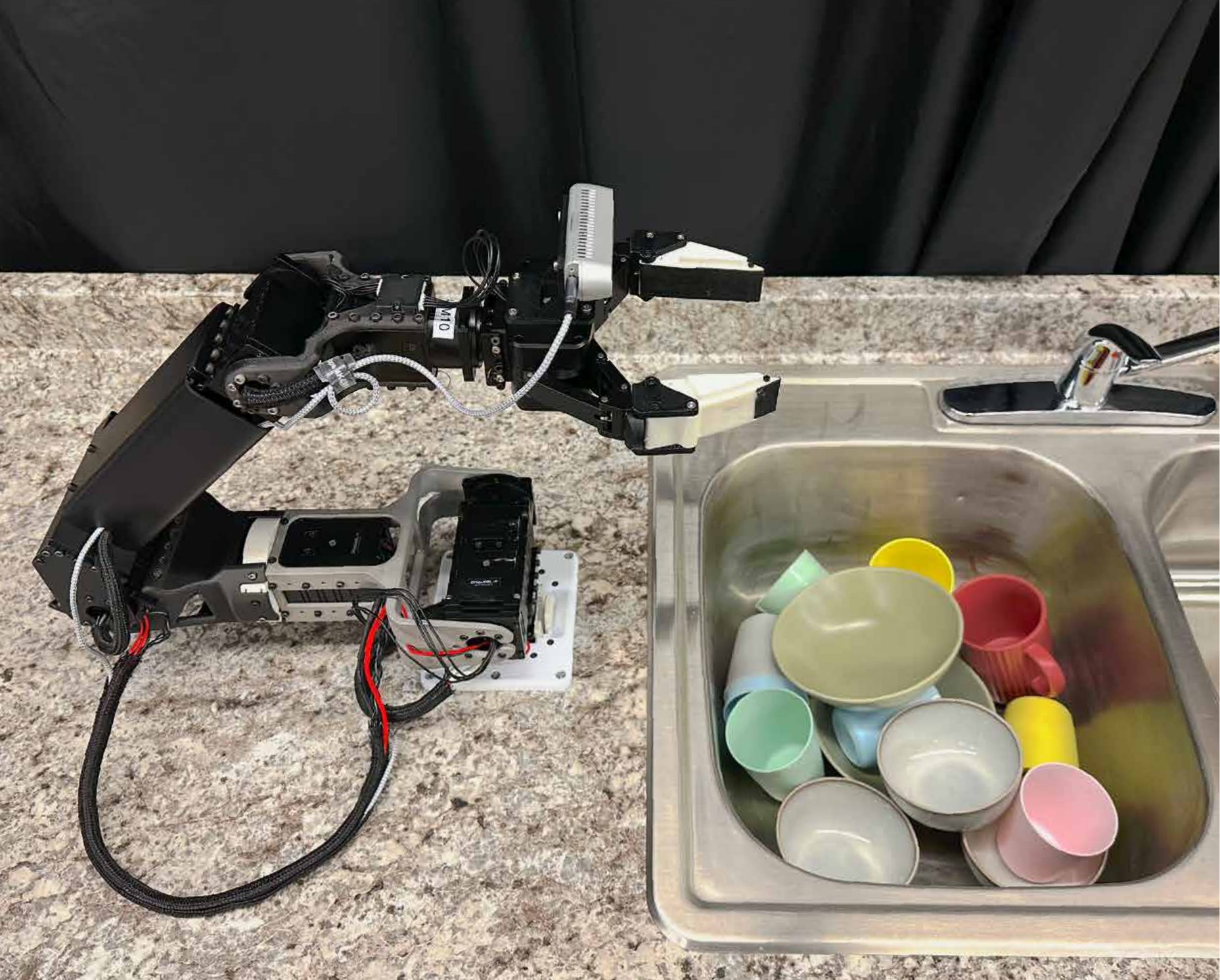}
    \label{fig:system2}
    }
    \subfigure[First Kitchen with Dishwasher]
    {%
    \includegraphics[width=0.9\columnwidth]{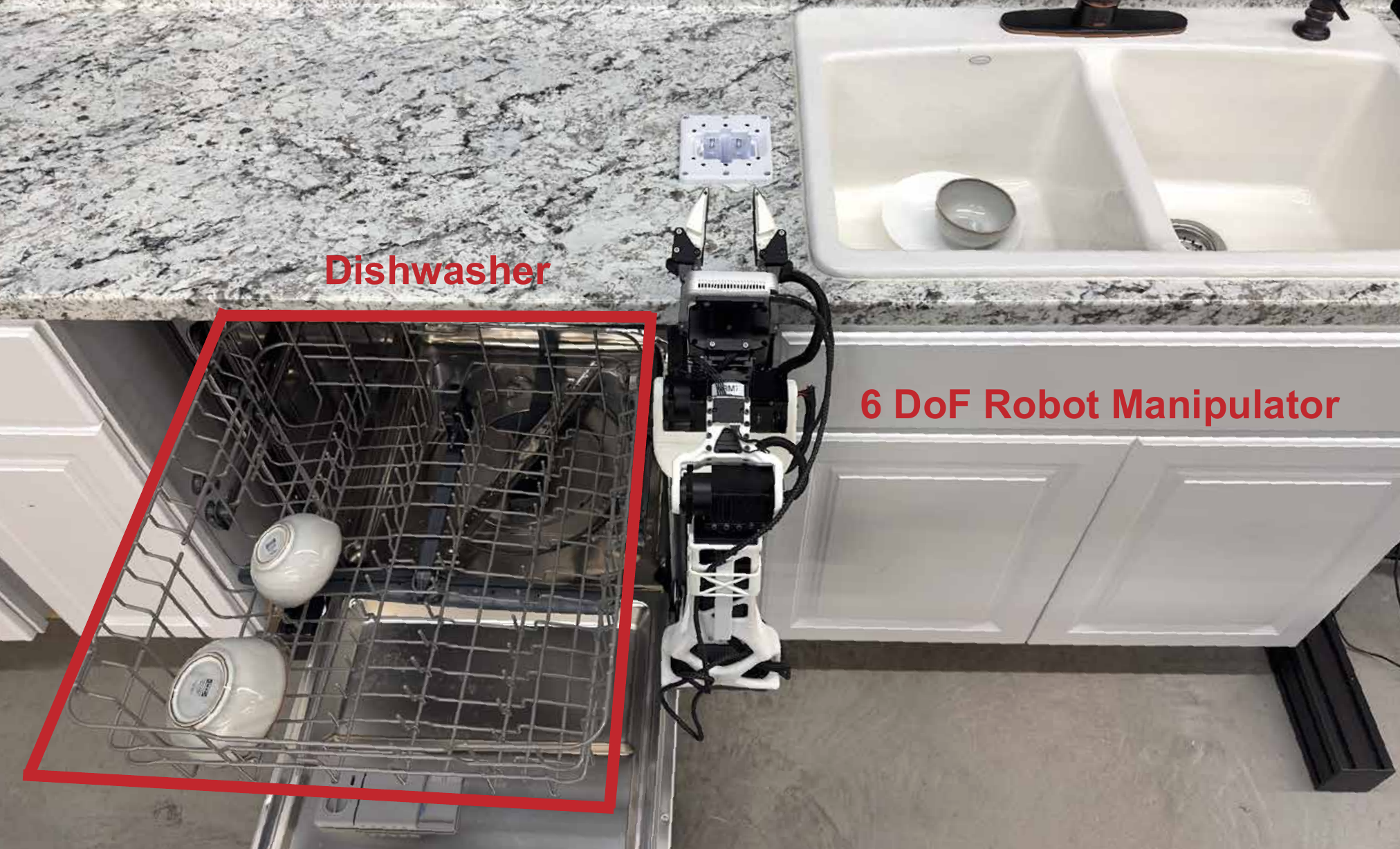}
    \label{fig:system3}
    }
    \end{minipage}

    \caption{Robot system configurations. Three configurations are prepared using two kitchen environments and two robotic arms.}
    \label{fig:robot_system}

\end{figure}

\section{Experimental Results}
\label{sec:exp}

In this section, we present the evaluation of the proposed perception pipeline using different combinations of visual and geometric foundation models. Based on this analysis, we identify the best-performing configuration. We further demonstrate its effectiveness through real-world full system experiments.

\subsection{Experiment Setup}

All experiments were conducted on a kitchen manipulation scenario using our custom robotic system \cite{kim2023papras} equipped with an RGB-D camera mounted on the end-effector. 
Hand-Eye calibration was performed using the Park–Martin method \cite{park1004handeye}. Pairwise $AX = XB$ consistency was evaluated across the retained calibration-pose pairs, yielding mean rotational and translational residuals of 0.614$\deg$ and 3.44 mm, respectively. The kinematic feasibility of each grasp candidate was evaluated using a Pinocchio \cite{carpentier2019pinocchio}-based numerical inverse-kinematics solver. Motion planning was performed using RRT-Connect \cite{kuffner2000rrtconnect}, while self-collision and environment collision were checked using \ac{FCL} \cite{pan2012fcl}.

We set up two kitchen environments to validate the generalizability of the presented pipeline, as shown in \figref{fig:robot_system}. 
Multi-view RGB-D observations of the sink workspace were acquired from three predefined end-effector observation poses. One RGB-D frame set was captured at each pose, yielding three frame sets in total.

For systematic evaluation, we instantiated different configurations of the perception pipeline by substituting the visual and geometric foundation models. The specific models used in our assessment are summarized in \tabref{tab:perception}. All learned networks were used without environment-specific training or fine-tuning.

\subsection{Dataset and Evaluation Metric}

To validate the pipeline, we built a custom dataset comprising 20 real-world
scenes captured with an RGB-D camera in two kitchen environments. The dataset
focuses on the sink and surrounding counter space, which constitute the
primary workspace for the dishware manipulation tasks considered in this
work. Across the scenes, we varied object arrangement, clutter level,
occlusion pattern, and sink material. Together, these factors reflect key
sources of scene variation in sink-area dishware handling, including the
spatial relationships among objects, their degree of visibility, and the
physical characteristics of the surrounding workspace. All scenes were manually annotated with object masks and 6D object poses following the procedure described in \cite{kim2024transpose}. The example sequences of this dataset are shown in the second row of \figref{fig:perception} and \figref{fig:mars_kitchen}.

Since most of the objects in our custom dataset are symmetrical, we adopt the Average Distance of model points for Indistinguishable views (ADI) as the evaluation metric. ADI measures the average distance between 3D model points transformed by the estimated pose and their closest counterparts on the model transformed by the ground-truth pose.

\begin{table*}[!t]
    \centering
    \small
    \setlength{\tabcolsep}{3.6pt}
    \renewcommand{\arraystretch}{1.18}
    \begin{adjustbox}{max width=0.99\textwidth}
        \begin{tabular}{@{}llcccc@{\hspace{0.8em}}cccc@{}}
            \toprule
            \multirow{2}{*}{\textbf{Visual model}}
                & \multirow{2}{*}{\textbf{Feature setting}}
                & \multicolumn{4}{c}{\textbf{LLMDet} + \textbf{SAMv2} }
                & \multicolumn{4}{c}{\textbf{GroundingDINO} + \textbf{SAMv2}} \\
            \cmidrule(lr){3-6}\cmidrule(l){7-10}
                & & \textbf{GeDI}
                & \textbf{FPFH} 
                & \textbf{BufferX} 
                & \shortstack{\textbf{Geo}\\\textbf{Transformer}}
                & \textbf{GeDI}
                & \textbf{FPFH} 
                & \textbf{BufferX} 
                & \shortstack{\textbf{Geo}\\\textbf{Transformer}} \\
            \midrule

            \multirow{2}{*}{DINOv2 }
                & without fusion & \textbf{88.19} & 84.26 & 85.16 & 76.92
                & 84.99 & 83.34 & 83.18 & 73.99 \\
                & with fusion    & 88.86 & 88.70 & 88.40 & \textbf{88.92}
                & 85.95 & 85.66 & 84.83 & 86.87 \\
            \addlinespace[2pt]
            \midrule

            \multirow{2}{*}{BEiT }
                & without fusion & 83.99 & 80.20 & 83.31 & 74.83
                & 82.44 & 79.57 & 81.42 & 74.42 \\
                & with fusion    & 85.18 & 84.20 & 83.84 & 84.45
                & 83.35 & 80.75 & 81.86 & 81.64 \\
            \addlinespace[2pt]
            \midrule

            \multirow{2}{*}{CLIP }
                & without fusion & 87.40 & 83.91 & 83.96 & 75.01
                & 85.25 & 83.02 & 82.47 & 76.32 \\
                & with fusion    & 88.44 & 87.77 & 88.34 & 87.11
                & 86.00 & 85.46 & 85.82 & 86.49 \\
            \bottomrule
        \end{tabular}
    \end{adjustbox}
    \caption{The quantitative result of the perception pipeline on the ADI metric for the custom kitchen-environment dataset. The tables report performance across different combinations of visual and geometric foundation models. \texttt{with fusion} denotes the use of fused 2D visual and 3D geometric features, whereas \texttt{without fusion} denotes the use of 3D geometric features only. All reported values are obtained without ICP refinement to highlight the effect of fused features.}
    \label{tab:perception}
\end{table*}

\begin{table}[!t]
    \centering
    \begin{adjustbox}{width=0.99\columnwidth}
    {
        \begin{tabular}{ccccc}
        \toprule
        \multirow{2}{*}{\textbf{method}}
            & \multicolumn{2}{c}{\textbf{Best Configuration}}
            & \multicolumn{2}{c}{\textbf{FoundationPose}
            } \\
        \cmidrule(lr){2-3}\cmidrule(l){4-5}

            & \textbf{with ICP}
            & \textbf{without ICP}
            & \textbf{with All GT Mask}
            & \textbf{with First GT Mask} \\

        \midrule
        ADI
            & \textbf{89.12}
            & 88.92
            & 87.32
            & Fail \\
        \bottomrule
        \end{tabular}
    }
    \end{adjustbox}

    \caption{Performance comparison with FoundationPose on the ADI metric for the custom kitchen-environment dataset. The best-performing configuration is LLMDet + SAMv2 + DINOv2 + GeoTransformer.}
    \label{tab:comparison}
\end{table}

\subsection{Analysis of The Perception Pipeline}

\subsubsection{Performance Analysis of Foundation Model Combinations}

\tabref{tab:perception} summarizes the quantitative results of the proposed perception pipeline on the custom dataset. We evaluated all combinations of two \ac{OVOD} methods, three visual foundation models, and four geometric foundation models. Results are reported both with and without the 2D–3D feature fusion strategy, and all values are obtained without ICP refinement in order to isolate the effect of fused features.

First, we compare two \ac{OVOD} methods, LLMDet and GroundingDINO. Both approaches provided reliable detection prompts for SAMv2, but LLMDet consistently yielded higher performance. By leveraging both image-level long captions and region-level short phrases during training, LLMDet appears to provide stronger generalization capability, even when using a broad prompt such as \texttt{"dishware"}. This advantage translates into more accurate and stable bounding boxes, which in turn provide higher-quality prompts for SAMv2, leading to more consistent instance masks across views.

When image and point features are used independently, the visual foundation models are primarily responsible for object candidate proposal, while the geometric foundation models are utilized for object registration. For the proposal, image features are extracted from object-centric crops obtained by masking out the background, resulting in localized object regions. In this context, effective candidate proposal requires features that can capture fine-grained locality within the object appearance. As shown in the non-fusion cases of \tabref{tab:perception}, while BEiT, based on masked image modeling, and CLIP, trained for image–text alignment, primarily provide strong global representations, DINOv2, trained with self-distillation, has been reported to produce more discriminative patch-level (fine-grained) features. These fine-grained features are particularly effective for candidate proposal in the perception pipeline, leading to the superior performance observed with DINOv2-based configurations.

In addition, for object registration within the perception pipeline, point-level features are extracted from each instance point cloud, and registration is performed based on point feature correspondences. Achieving strong registration performance requires descriptors that can produce highly discriminative representations in local 3D regions. As reported in the non-fusion cases of \tabref{tab:perception}, GeDi provides the most effective local descriptors among the evaluated methods, as it is trained with a local reference frame. This property enables robust registration even under cluttered and partially occluded conditions. In contrast, GeoTransformer emphasizes non-local contextual information for point cloud registration, which leads to inferior performance when applied within our perception pipeline, where fine-grained local correspondences are critical.

When image and point features are fused, the performance of the pipeline improves consistently across all configurations, as shown in the fusion cases of \tabref{tab:perception}. This confirms that the two modalities provide complementary information: image features contribute semantic and appearance cues, while point features encode structural details. The fusion strategy integrates these representations, resulting in more robust discrimination even under clutter and partial occlusions.

The most significant improvement is observed with GeoTransformer. Although its standalone performance is relatively low due to its reliance on global geometric context, when fused with discriminative image features, its global modeling capability is strongly reinforced, leading to the highest overall accuracy. In contrast, GeDi already provides strong local descriptors in the no-fusion setting, and thus the relative gain from fusion is smaller. FPFH and BufferX also benefit substantially from fusion, as the semantic information from the visual foundation model compensates for their weaker discriminative power.

Among all tested configurations, the combination of DINOv2 as the visual foundation model, LLMDet as the \ac{OVOD}, and GeoTransformer as the geometric foundation model achieved the best performance, reaching 88.92\% ADI accuracy without ICP refinement. This result highlights the synergy between fine-grained semantic cues from DINOv2 and the global geometric modeling capability of GeoTransformer, which together yield the most effective configuration of the proposed perception pipeline.

\begin{figure*}[!t]
	\centering
	\includegraphics[width=0.99\textwidth]{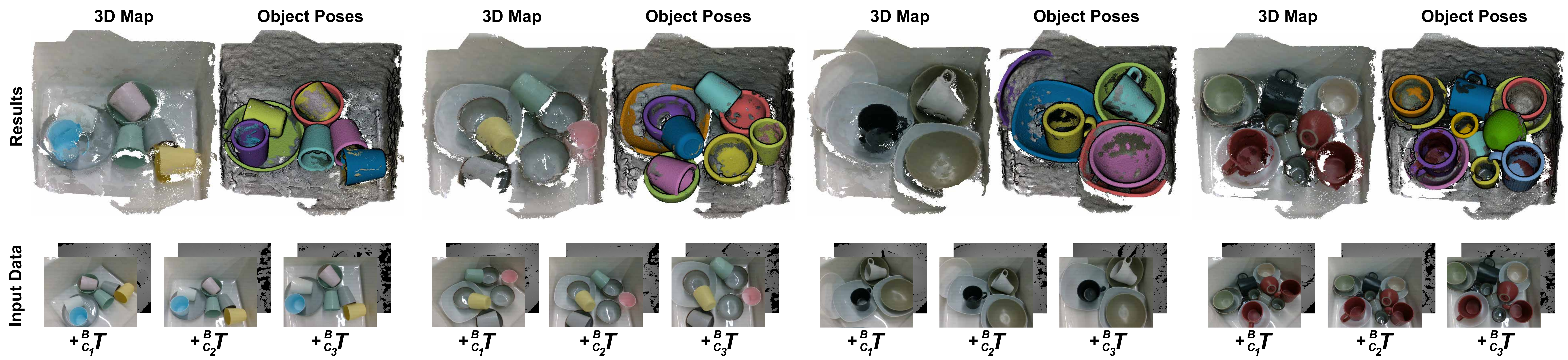}
  	\caption{The qualitative evaluation of the best combination model. The first row illustrates the reconstructed 3D map and the estimated object poses using 3D object models. The second row shows samples from our custom kitchen-environment dataset.}
	\label{fig:perception}
\end{figure*}

\begin{figure}[!t]
	\centering
	\includegraphics[width=0.99\columnwidth]{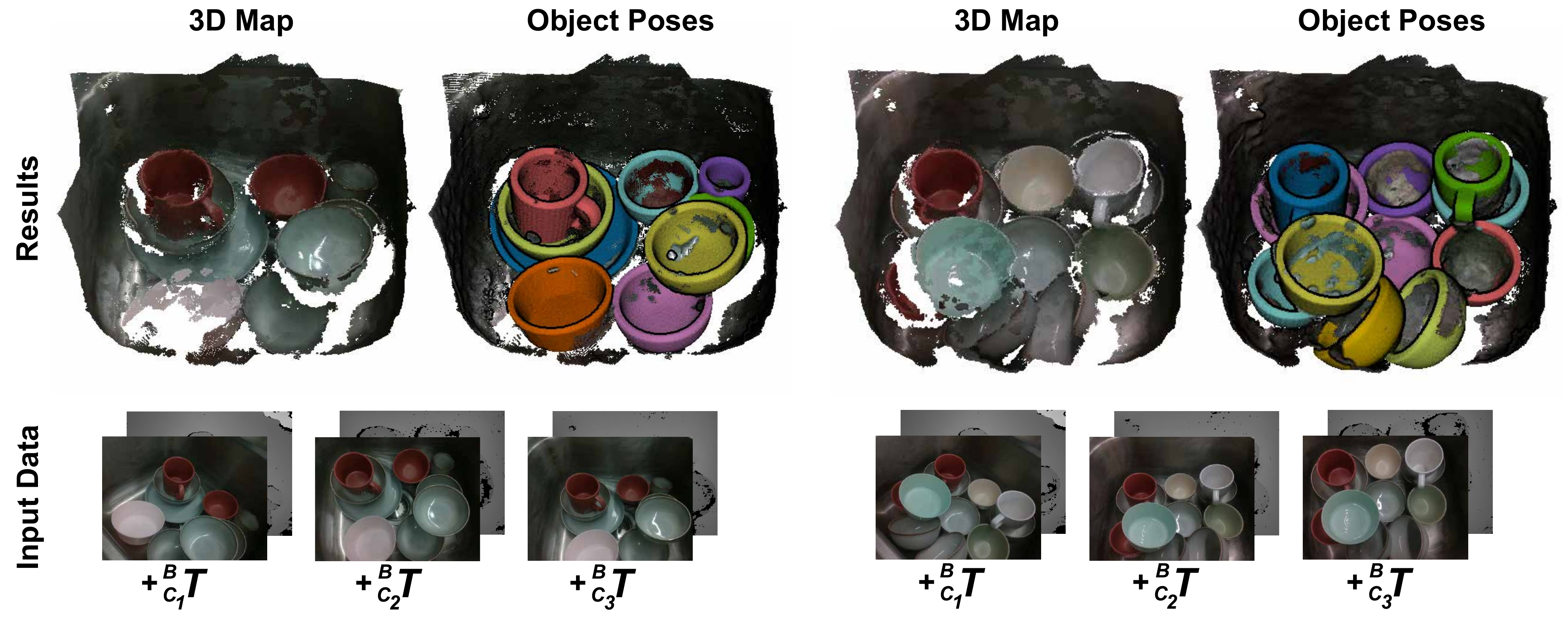}
  	\caption{The qualitative evaluation of the best model on the second kitchen environment.}
	\label{fig:mars_kitchen}
\end{figure}

To further refine the estimated object poses, we apply ICP refinement to the best-performing configuration of the perception pipeline. As reported in \tabref{tab:comparison}, ICP refinement yields a modest improvement in accuracy. We also compare this configuration against FoundationPose. Since our custom kitchen-environment dataset does not provide fully sequential input images, FoundationPose is unable to reliably track the target objects. To address this limitation, we supply the ground-truth mask for each target object, which represents an ideal condition that is not available in practical deployment scenarios. Even under this favorable condition, our best-performing configuration achieves higher accuracy than FoundationPose. Our modular pipeline not only achieves higher accuracy but also provides flexibility to adapt to different computational constraints and emerging foundation models. The qualitative results on the custom kitchen-environment dataset are presented in \figref{fig:perception}, while the adaptability of the perception pipeline is further demonstrated on a different kitchen environment, as shown in \figref{fig:mars_kitchen}.

%FIGURE
%FIGURE

%FIGURE
%FIGURE

\subsubsection{Sensitivity Analysis of Key Hyperparameters}

\begin{figure*}[!t]
  \centering
  \subfigure[]{%
    \includegraphics[width=0.32\textwidth]{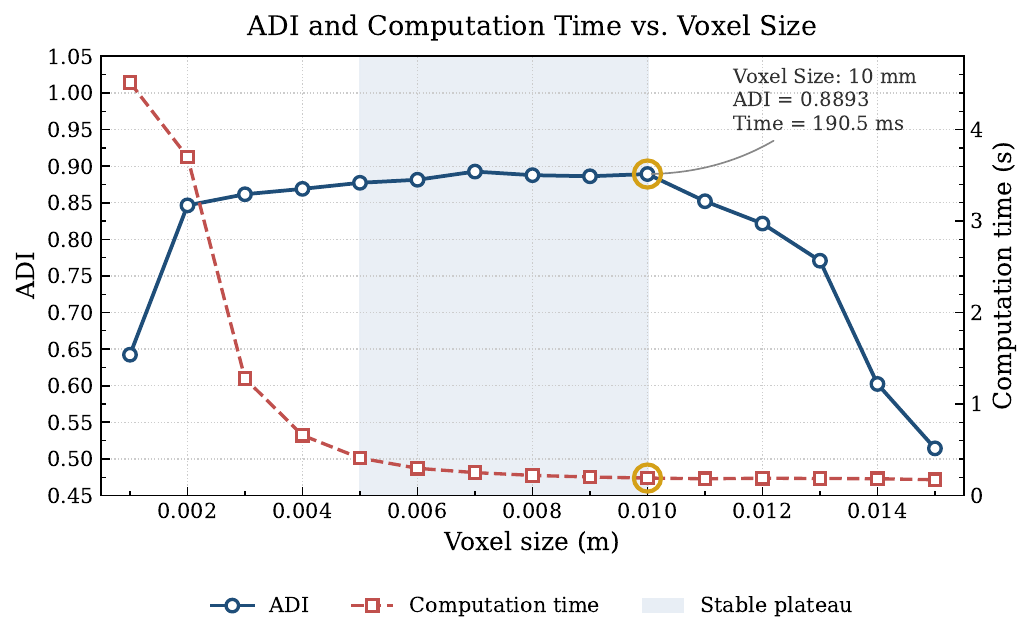}    
    \label{fig:adi_vs_voxel}
  }%
  \subfigure[]{%
    \includegraphics[width=0.32\textwidth]{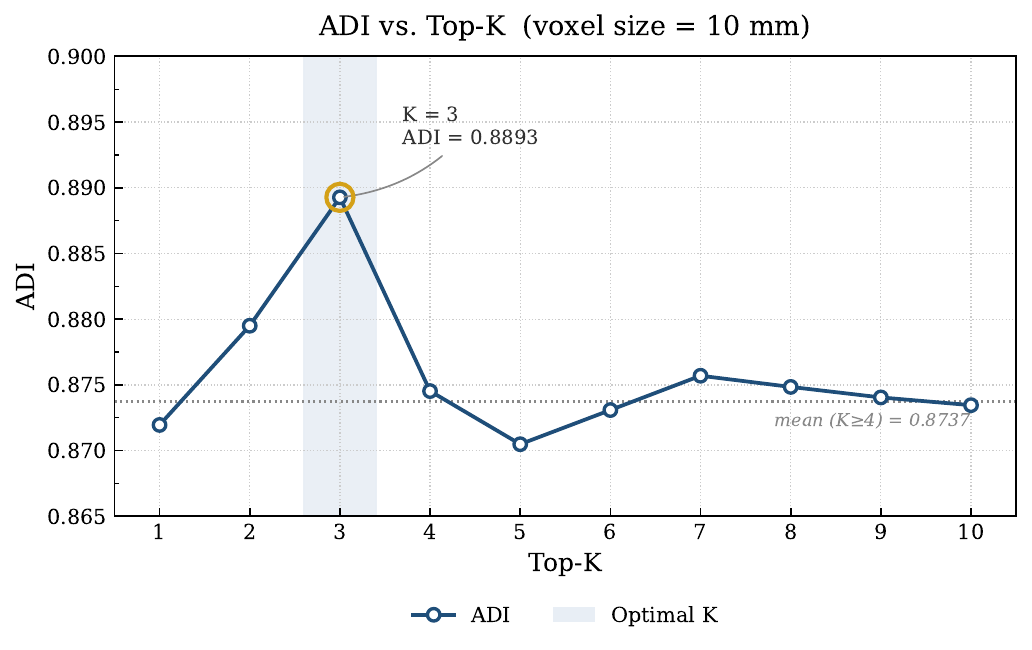}
    \label{fig:adi_vs_topk}
  }
  \subfigure[]{%
    \includegraphics[width=0.32\textwidth]{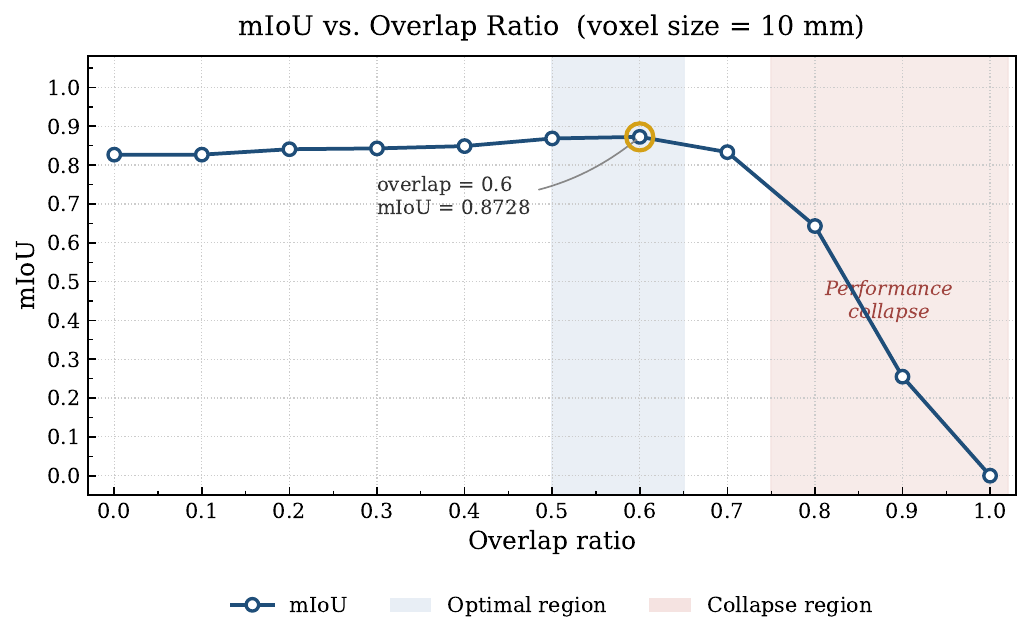}
    \label{fig:adi_vs_overlap}
  }  
  \caption{Performance variation with respect to \subref{fig:adi_vs_voxel} voxel size, \subref{fig:adi_vs_topk} Top-K, and \subref{fig:adi_vs_overlap} overlap ratio. The highlighted regions indicate the optimal operating ranges, with optima found at voxel size = 10 mm, K = 3, and overlap ratio = 0.6.}  
  \label{fig:ablation}
\end{figure*}

Beyond the choice of foundation models, the perception pipeline involves several hyperparameters that influence its accuracy. We analyze three of them with the best-performing configuration (LLMDet + SAMv2 + DINOv2 + GeoTransformer): the voxel size used for point cloud processing, the number of top-$K$ candidates retained during object candidate proposal, and the IoU threshold used for cross-view mask association in multi-view instance segmentation. \figref{fig:ablation} summarizes the results.

The voxel size determines the trade-off between spatial resolution and computational cost. As shown in \figref{fig:adi_vs_voxel}, ADI rises sharply from 0.6423 at 1\,mm to a stable plateau between 5\,mm and 10\,mm, remaining within the range of 0.8772–0.8926. Beyond 11\,mm, performance degrades rapidly as the resolution becomes insufficient to preserve discriminative geometric details. Computation time decreases monotonically with increasing voxel size and saturates around 10\,mm. We therefore adopt 10\,mm as the operating voxel size, achieving an ADI of 0.8892 at 190.5\,ms per object while providing the best balance between accuracy and runtime.

The Top-$K$ parameter determines how many candidate CAD models are retained after cosine-similarity-based filtering for subsequent geometric registration. As shown in \figref{fig:adi_vs_topk}, ADI peaks at $K=3$. When $K$ is too small, the correct candidate may be pruned before registration. In contrast, larger $K$ values introduce visually similar but incorrect candidates, which can occasionally produce deceptively high overlap scores during registration. We therefore set $K=3$, as it provides sufficient recall while effectively suppressing such distractors, and use this setting in all subsequent experiments.

The IoU threshold controls the strictness of cross-view mask association in multi-view instance segmentation. Masks from different views are assigned to the same object if their projected overlap exceeds this threshold. As shown in \figref{fig:adi_vs_overlap}, mIoU increases gradually as the threshold rises from 0.0 to 0.6, reaching a peak of 0.873. Then it drops sharply beyond 0.7. Low thresholds tend to merge masks from different objects, reducing instance purity. In contrast, overly high thresholds reject valid correspondences, as depth noise and viewpoint variation prevent perfect projective overlap even for the same object. We therefore set the threshold to 0.6, which is sufficiently strict to avoid over-merging while remaining tolerant enough to preserve consistent instances across views.

\subsection{Real-World Full System Demo}

\begin{figure*}[!t]
	\centering
	\includegraphics[width=0.99\textwidth]{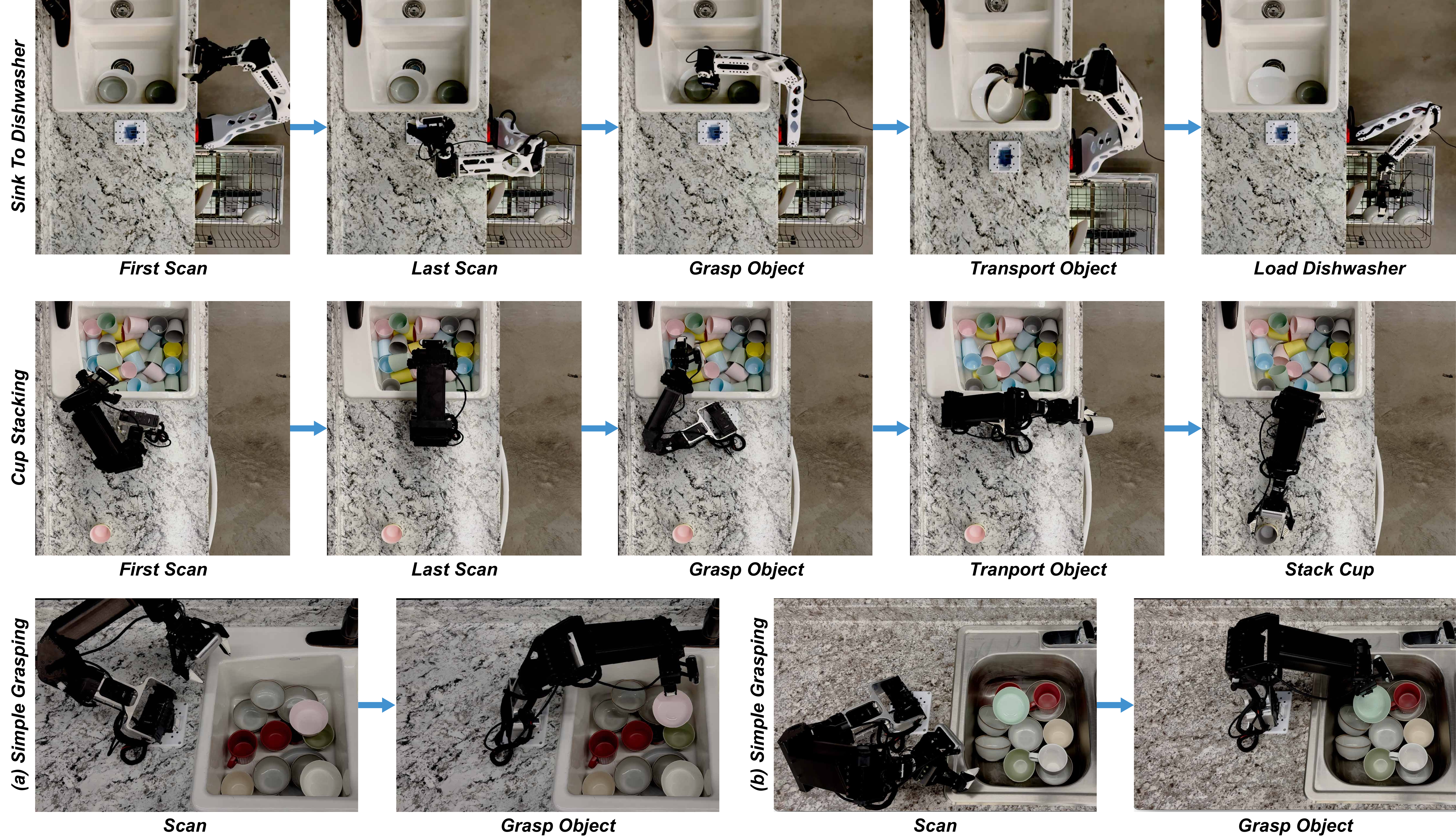}
  	\caption{Tasks performed to demonstrate the best model in a real kitchen environment.}
	\label{fig:task}
\end{figure*}

To further validate the perception pipeline, we deploy the best-performing configuration on a physical robotic system performing kitchen manipulation tasks. \figref{fig:task} shows three representative demonstrations: transferring dishware from a sink to a dishwasher, cup stacking on a counter, and simple grasping of dishware from a sink.

In the sink-to-dishwasher scenario, the robot system reliably detects dishware, estimates poses, and executes collision-free grasps to transport objects into the dishwasher. In the cup-stacking task, the robot demonstrated precise placement and alignment, confirming the accuracy of the estimated poses. In the simple grasping task, the robot maintains robustness across the different kitchen environments. These demonstrations showcase the scalability and practicality of the pipeline in real-world household manipulation, without any environment-specific retraining. Additional experimental results are available in the supplementary video materials.

\subsubsection{Quantitative evaluation}
To quantitatively evaluate the reliability of the proposed pipeline, we report the success rates of the simple grasping and cup stacking tasks in \tabref{tab:success_rate}. Across 296 trials, the pipeline achieves 259 successful executions, yielding success rates above 87\% for all tasks. The consistently high performance across different tasks demonstrates the practical reliability and robustness of the proposed pipeline.

\subsubsection{Failure analysis}

Across all tasks, the majority of failures originate from gripper slip during grasp execution. We adopt a compliant gripper inspired by the Fin-Ray Gripper \cite{crooks2016fin} to enable softer and more adaptive grasping. However, its compliant structure limits the maximum grasp force, occasionally causing objects to slip from the gripper during transport. In addition to these physical interaction failures, rare failures also arise from inaccurate detections that lead the robot to grasp incorrect locations.

In the cup stacking task, the sink environment contains heavily cluttered and densely stacked cups, introducing an additional failure mode. During gripper closure, contact with neighboring cups can unintentionally displace the target cup, causing it to deviate from the estimated grasp pose and resulting in grasp failure. These observations indicate that the remaining failures mainly arise from physical interaction challenges. Representative failure cases are provided on the project page.

\subsubsection{Runtime analysis}

We further analyze the runtime of the proposed pipeline on the real-world tasks using a workstation equipped with an AMD Ryzen9-7950X and an NVIDIA RTX 4090. Constructing the 3D map and performing instance segmentation from three scanned images requires 1.9 $s$ on average. Subsequently, pose and grasp estimation require 190 $ms$ per object and scale linearly with the number of detected objects. In most cases, the number of detected objects rarely exceeds 20, resulting in approximately 3.8 $s$ for the pose estimation stage in the worst case. Finally, motion planning for the robot manipulator takes approximately 2 $s$ on average. Consequently, the overall runtime is up to 7.7 $s$ in the worst case and gradually decreases as the task progresses, since the number of remaining objects in the workspace decreases after each successful manipulation.

%FIGURE
%FIGURE

\begin{table}[!t]
    \centering
    \small
    \setlength{\tabcolsep}{5pt}
    \renewcommand{\arraystretch}{1.18}

    \begin{adjustbox}{max width=0.99\columnwidth}
        \begin{tabular}{@{}lcccc@{}}
            \toprule
            \textbf{Task}
                & \textbf{First Kitchen}
                & \textbf{Second Kitchen}
                & \textbf{Cup Stacking}
                & \textbf{All} \\
            \midrule

            Number of Trials
                & 90 & 74 & 132 & 296 \\

            Number of Successes
                & 81 & 67 & 111 & 259 \\

            \midrule

            Success Rate (\%)
                & 90.00
                & 90.54
                & 84.09
                & \textbf{87.50} \\

            \bottomrule
        \end{tabular}
    \end{adjustbox}

    \caption{Success rate of the best-performing configuration in real-world tasks.}
    \label{tab:success_rate}
\end{table}

\section{Conclusion and Future Works}
\label{sec:conclusion}

In this paper, we presented a modular perception pipeline for robotic manipulation in kitchen environments. By integrating open-vocabulary object detection, multi-view segmentation, instance-aware 3D reconstruction, and 2D–3D feature fusion, the pipeline enables robust 6D pose estimation and grasp planning without environment-specific retraining. Through systematic evaluation of different combinations, we identified the best-performing configuration, which maintained strong accuracy even under cluttered and occluded conditions. Real-world demonstrations further validated this configuration, showing that it can be directly deployed on physical robots to execute tasks such as sink-to-dishwasher transfer and cup stacking. These results confirm the scalability and adaptability of the proposed approach and highlight its potential as a practical framework for household robotics.

The modular architecture empowers practitioners to tailor model combinations to their specific constraints: utilizing streamlined configurations for efficiency or sophisticated ensembles for high-difficulty scenarios. This flexibility facilitates the optimization of computational resources without sacrificing the advantages of our no-task-specific-finetuning design.

Despite these promising results, several limitations remain, which suggest directions for future research. First, grasping failures were occasionally observed due to the constraints of the parallel-jaw gripper, indicating the need for more versatile end-effector designs or adaptive grasp planners that can handle diverse object geometries. Second, some scenarios demanded more complex manipulation skills, such as extracting tightly stacked or occluded objects, which require multi-step or non-prehensile strategies beyond the current framework. Extending the system to incorporate such advanced manipulation primitives will be essential for addressing real-world variability. Third, while the current planner ensures collision-free execution, it lacks optimization for path length, smoothness, and naturalness, often resulting in jerky or circuitous trajectories in cluttered environments. Future work will address this by incorporating a learned trajectory generator conditioned on 6D-pose estimates to achieve more natural motion and improved performance in cluttered environments. Fourth, with sufficient pose-labeled data, learned fusion modules could further improve visual-geometric feature representations beyond the simple concatenation used in this work. Finally, the proposed pipeline relies on CAD models for instance-level registration and grasp configuration, which is appropriate for the inventory-based kitchen scope of this work but restricts broader applicability. Future work could leverage multi-view reconstruction methods to acquire object models through a brief on-boarding scan when CAD is unavailable.

\balance
% \small
\bibliographystyle{IEEEtranN}
\bibliography{string-short,references}

\end{document}